\pdfoutput=1
\documentclass[runningheads]{llncs}
\usepackage[T1]{fontenc}
\usepackage{graphicx}
\usepackage{bbding}
\usepackage{amsmath} 
\usepackage{amssymb}
\usepackage{subcaption}
\usepackage{tabularx}
\usepackage{multirow} 
\usepackage{booktabs}
\usepackage{siunitx}
\usepackage{caption}
\newcolumntype{P}[2]{S[table-format=#1.#2]}

\begin{document}
\title{S$^2$tory: Story Spine Distillation for Movie Script Summarization}

\author{Mingzhe Lu\inst{1,2} \and
	Yanbing Liu\inst{1,2} \and
	Qihao Wang\inst{1,2} \and
	Jiarui Zhang\inst{1,2} \and
	Jiayue Wu\inst{1,2} \and
	Yue Hu\inst{1,2} \and
	Yunpeng Li\Envelope\thanks{Corresponding author: liyunpeng@iie.ac.cn} \inst{1,2}  \and
	Yangyan Xu\inst{3}}

\institute{Institute of Information Engineering, Chinese Academy of Sciences, Beijing, China \and
	School of Cyber Security, University of Chinese Academy of Sciences, Beijing, China \and
	HiThink Research, Hangzhou, China}



%
\maketitle              
\begin{abstract}
Movie scripts pose a fundamental challenge for automatic summarization due to their non-linear, cross-cut narrative structure, which makes surface-level saliency methods ineffective at preserving core story progression. To address this, we introduce S$^2$tory (Story Spine Distillation), a narratology-grounded framework that leverages character development trajectories to identify plot nuclei, the essential events that drive the narrative forward, while filtering out peripheral satellite events that merely enrich atmosphere or emotion. Our Narrative Expert Agent (NEAgent) performs theory-constrained reasoning, whose distilled knowledge conditions a small model to identify plot nuclei. Another model then uses these plot nuclei to generate the summary. Experiments on the MovieSum dataset demonstrate state-of-the-art semantic fidelity at approximately 3.5$\times$ compression, and zero-shot evaluation on BookSum confirms strong out-of-domain generalization. Human evaluation further validates that narratological theory provides an indispensable foundation for modeling complex, non-linear narratives.

\end{abstract}
\section{Introduction}
Large language models (LLMs) have achieved remarkable progress in text understanding. However, their performance declines on long-form and structurally complex narratives, especially in maintaining the core storyline. This limitation becomes particularly evident in movie script summarization, which poses unique challenges~\cite{saxena2024moviesum} beyond traditional text summarization.

Long-form summarization typically adopts a two-stage process: shortening the source, then generating the summary~\cite{mei2025survey}. Among recent approaches in movie script summarization, MENSA~\cite{saxena2024select} selects scenes based on saliency estimation, while DiscoGraMS~\cite{chitale2025discograms} constructs a graph over characters and dialogues to model cross-scene coherence. Despite their effectiveness, these approaches remain largely data-driven and rely on lexical or shallow structural patterns.

Movie scripts are shaped by symbolic narrative structures, a foundational element of screenwriting, and effective summarization must recover this underlying logic. Capturing these structures requires reasoning about how events function within the story rather than how they appear in text. This gap motivates a model that integrates symbolic representation with neural abstraction to reconstruct the narrative backbone of a movie, an aspect largely overlooked in previous work.

Inspired by Barthesian narrative theory~\cite{barthes1975introduction}, we posit that movie script summarization requires explicit modeling of narrative structure to overcome the limitations of saliency-based approaches. Accordingly, we propose S$^2$tory, a framework that identifies plot nuclei as the essential events forming the core narrative thread and uses them as the foundation for high-fidelity summarization.

In S$^2$tory, a Narrative Expert Agent (NEAgent) follows narratological principles to identify plot nuclei essential to narrative integrity. The NEAgent analyzes the screenplay through character-arc trajectories, using character development to determine the indispensability of plot events. A student model distills this reasoning, and a fine-tuned summarizer performs nuclei-conditioned summarization.

In summary, our main contributions are as follows:
\begin{itemize}
	\item We propose S$^{2}$tory, a narratologically-grounded framework that includes a NEAgent for identifying plot nuclei through character-arc trajectories.
	\item Within S$^{2}$tory, we introduce a distillation step where a student model learns the reasoning of the NEAgent, and a fine-tuned summarizer performs nuclei-conditioned summarization.
	\item We achieve state-of-the-art performance on the MovieSum benchmark, and demonstrate out-of-distribution generalization on BookSum through automated validation, human evaluation, and case studies.
\end{itemize}

\section{Related Work}
\subsection{Computational Narratology}
Computational narratology, rooted in classical narratology~\cite{chatman1978story}, aims to operationalize how stories produce meaning. Early approaches focused on symbolic representations, such as story grammars~\cite{rumelhart1975notes}, plot units~\cite{lehnert1981plot}, and character-centered scripts~\cite{schank2013scripts}, yielding interpretable but domain-limited frameworks. The field subsequently evolved into LLM-driven approaches, which can implicitly capture narrative patterns in text~\cite{huang2024ecr}. However, despite their proficiency in narrative processing, LLMs lack a deep understanding of underlying relational structures and narrative functions~\cite{brahman2022modeling}. Recent studies have re-grounded narrative modeling in theory via turning points and events~\cite{jiayang2024eventground}, yet they still fail to capture the Barthesian~\cite{barthes1975introduction} distinction between structural necessity and surface prominence.

\subsection{Long-Text Summarization}
Long-text summarization poses distinct challenges for LLMs. Transformer architectures face quadratic complexity, leading to long-context variants such as Longformer~\cite{beltagy2020longformer}, BigBird~\cite{zaheer2020big}, and LongT5~\cite{guo2022longt5}. Although these models extend context windows, they only partially capture global structure and still treat narratives as flat token sequences, missing the multi-level organization~\cite{beltagy2020longformer} inherent in human storytelling. In movie script summarization, datasets like SummScreen~\cite{chen2022summscreen} and MovieSum~\cite{saxena2024moviesum} highlight the need to model cross-scene coherence and characterdriven progression. Their coherence arises from causal and thematic continuity rather than textual adjacency. Building on this, structure-aware models such as DiscoGraMS~\cite{chitale2025discograms} and ScreenWriter~\cite{mahon2024screenwriter} incorporate discourse or character graphs but remain limited to surface cues-modeling who interacts with whom, but not why those interactions matter narratively.

\section{Narrative Theoretical Formulation}
\subsection{Narratological Foundation}
S$^{2}$tory is grounded in Barthes' narratological distinction between nuclei and satellites~\cite{barthes1975introduction}. In a narrative, nuclei are essential events that drive the core progression of the story, while satellites are auxiliary elements that enrich atmosphere or emotion without altering the narrative trajectory.

\subsection{Narrative World Modeling}
In computational narrative modeling, we represent a story world as a quadruple $(\mathcal{C},\mathcal{E},\mathcal{S},\mathcal{R})$ , where $\mathcal{C}$ is the set of characters, $\mathcal{E}$ is the set of events, $\mathcal{S}$ is the space of character states, and $\mathcal{R}\subseteq\mathcal{S}\times\mathcal{S}$ is the set of state transition relations.

Each transition r: s$_{t}$ $\rightarrow$ $s_{t+1}$ is an element of $\mathcal{R}$, capturing a character's developmental trajectory. Based on the causal origin of r, we classify it into two types: intrinsic and extrinsic transitions. Intrinsic transitions arise from internal processes such as reflection, cognitive reevaluation, or emotional realization. Extrinsic transitions, by contrast, are triggered by external events $e \in \mathcal{E}$ , representing environmental or social influences that compel a character to adapt.

Our approach focuses on identifying events that directly shape character development. We define a subset $e^{\rightarrow s}\subseteq \mathcal{E}$ as the collection of events that causally induce a state transition in at least one character:
\begin{equation}
e^{\rightarrow s} = \left\{e \in \mathcal{E} \mid \exists r: s_{t} \rightarrow s_{t+1} \in \mathcal{R}, \operatorname{Dep}(r, e) \wedge s_{t + 1} \neq s_{t} \right\},
\end{equation}
where $\mathrm{Dep} (r, e)$ indicates that the transition r is causally dependent on event e and $s_{t+1}\neq s_{t}$ ensures that the event results in an actual change in the character's state. This definition operationalizes narrative significance:
\begin{itemize}
	\item An event is considered narratively relevant if it leads to a detectable change in a character's state.
	\item Conversely, an event whose removal blocks character growth is crucial, marking a plot turning point.
\end{itemize}

\subsection{Modeling Character Dynamics}
Each character $c\in \mathcal{C}$ maintains a time-dependent attribute set $A_{c}^{(t)}=\{(k_{j},v_{j})\}$ representing evolving properties such as identity, goals, or affiliations.

A state transition $r: s_{t}\rightarrow s_{t+1}$ for character c is said to be event-induced if there exists $e\in \mathcal{E}$ such that $\operatorname{Dep} (r, e)$ and $A_{c}^{(t+1)} \neq A_{c}^{(t)}$ . We classify each such transition by its update type: $\tau(r)\in \{ \ominus, \oplus \}$ , where $\oplus$ denotes an \textit{increment} (addition of a new attribute) and $\ominus$ a \textit{modification} (replacement of an existing attribute). The state difference associated with r is:
\begin{equation}
\Delta(r) = \left(S^{+}, S^{-}\right), \quad A_{c}^{(t+1)} = \left(A_{c}^{(t)} \setminus S^{-}\right) \cup S^{+},
\end{equation}
with $S^{+}=A_{c}^{(t+1)}\backslash A_{c}^{(t)}$ and $S^{-}=A_{c}^{(t)}\backslash A_{c}^{(t+1)}$.


For example, being knighted is an \textit{increment}$(\oplus)$, adding a role without erasing history; conversely, moving cities is a \textit{modification}$(\ominus)$, replacing the old location.

\subsection{Narrative Nuclei Reasoning}
This module implements the nucleus-satellite distinction via counterfactual reasoning over character state trajectories, as formalized in the previous subsections.

\textit{Narrative context}. To analyze a scene $\mathcal{C}_{k}$ , the model uses the processed prior scenes, the structured character states, and the naturalized current scene $\tilde{\mathcal{C}}_{k}$ Based on this context, each sentence $u_{k,i}$ in $\tilde{\mathcal{C}}_{k}$ is evaluated to see if removing it would break character development.

\textit{Reasoning logic}. Define Cont $(\mathcal{M}_{k},\cdot)$ as a predicate that returns 1 if all character state trajectories remain structurally continuous. The nucleus-satellite label is assigned by:

\begin{equation}
	\kappa \left(u_{k,i}\right) = \left\{ \begin{array}{l l} 1, & \text {if} \neg \operatorname {Cont} \left(\mathcal {M}_{k}, \left(\tilde {\mathcal{C}}_{k} \setminus \{u_{k,i}\} \right) \cup B_{< k}\right), \\ 0, & \text{otherwise}. \end{array} \right.
\end{equation}
here, $\kappa(u_{k,i})=1$ indicates a \textit{nucleus}, and 0 a \textit{satellite}. A soft prior from the transition type $\tau(r)\in\{\oplus,\ominus\}$ can modulate the sensitivity of Cont($\cdot$), but does not override the continuity test. For instance, given two candidate units in $\tilde{C}_{k}$:

(1). $u_{k,1}$ : "Leon claims to be a deserter; mutual distrust is established."

(2). $u_{k,2}$ : "On the road, they encounter bandits, a plague village, and a rebel checkpoint."

Given $\mathcal{M}_{k}$ and $B_{<k}$ , the system tests whether deletion breaks any character trajectory. For $u_{k,1}$, removal disrupts Leon's trust evolution, causing a critical break in narrative coherence. Consequently, it is classified as a nucleus $u_{k,1}^{+}$. In contrast, removing $u_{k,2}$ leaves all state transitions intact and coherence preserved, marking it as a satellite $u_{k,2}^{-}$.

This reasoning process defines narrative indispensability in functional terms, where an event is considered a nucleus if and only if its removal breaks the continuity of character development within the evolving symbolic world.

\begin{figure}[t]
	\centering
	\includegraphics[width=0.95\textwidth]{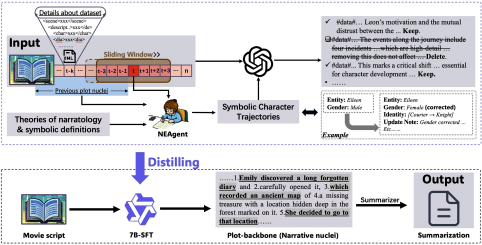}
	\caption{The S$^2$tory framework leverages character-guided reasoning to identify plot nuclei and filter satellites, thereby distilling expert knowledge to effectively condition abstractive summarization.} 
	\label{fig1}
\end{figure}

\section{Methodology}
In S$^{2}$tory, NEAgent models characters, infers plot nuclei under narrative constraints, and distills this reasoning into a nuclei-conditioned backbone for summarization. The overall process is illustrated in Fig~\ref{fig1}.

\subsection{NEAgent}
Implementing narratological reasoning remains challenging: symbolic models lack openness, while neural models lack narrative data. We address this by designing NEAgent, an In-Context Learning (ICL) agent that integrates narratological principles with dynamic character modeling.

At step \textit{t}, NEAgent processes the current narrative segment story$_{t}$ through ICL prompts that operationalize the narrative world model (Sec. 3.2) and character dynamics formalism (Sec. 3.3). The agent combines story$_{t}$ with a rolling memory $M_{t-1}$, which stores evolving character states $A_{c}^{(t)}$ and event-character dependencies $\mathrm{Dep}(r,e)$ (Sec. 3.4). This creates a contextualized narrative representation. Guided by the prompts, NEAgent evaluates each narrative unit $u_{k,i}$ using the nucleus-satellite classification rule in Eq.(2), determining whether its removal would disrupt character trajectory continuity. The prompts explicitly encode narratological principles as actionable instructions, ensuring alignment with the theoretical framework.

\subsection{Distillation and Summarization}
NEAgent offers detailed reasoning at high token cost, hindering large-scale use. To make this reasoning scalable, we record its analytical process as an experience context and construct a structured dataset as follows:
\begin{equation}
	\mathcal{D}_{\mathrm{distill}} = \left\{\left(x_{i}, r_{i}, B_{i}\right) \right\}_{i = 1}^{N},
\end{equation}
where $x_{i}$ is the naturalized narrative text of a scene, $r_{i}$ denotes the symbolic reasoning trace generated by the NEAgent, and $B_{i}$ is the corresponding set of nuclei identified through trajectory-guided counterfactual reasoning.

The reasoning trace $r_{i}$ encodes how the agent tracks character-state transitions, redefines goals, and evaluates whether removing an event would disrupt continuity within the evolving narrative structure.

The neural inducer is initialized from a 7B language model and fine-tuned on $\mathcal{D}_{\mathrm{distill}}$. It learns a mapping:
\begin{equation}
	f_{\theta}: \left(x_{\mathrm{shot}}^{n}, r_{\mathrm {shot}}^{n}, b_{\mathrm {shot}}^{n}, x_{i}\right)\rightarrow \left(r_{i}, B_{i}\right),
\end{equation}
where each training instance includes few-shot examples $(x_{\mathrm{shot}}^{n}, r_{\mathrm{shot}}^{n}, b_{\mathrm{shot}}^{n})$ and the current input $x_{i}$, to generate the corresponding reasoning trace $r_{i}$ and predicted nuclei set $B_{i}$ .

Finally, each backbone $B_{i}$ predicted by the distilled model is paired with its corresponding gold summary $y_{i}$ from the dataset, forming a new training set

\begin{equation}
	\mathcal{D}_{\mathrm{sum}} = \left\{\left(B_{i}, y_{i}\right) \right\}_{i=1}^{M}.
\end{equation}

The summarization model is then trained on $\mathcal{D}_{\mathrm{sum}}$ , where the model learns to generate reference summaries conditioned on the distilled backbones.

\section{Experiment}
\subsection{Setup and Implementation Details}
\paragraph{NEAgent.} The NEAgent was implemented in LangGraph with GPT-4o as the reasoning engine, operating deterministically with temperature set to 0.0. See Appendix~\ref{app:prompt_detail} for a comprehensive description of the prompts.

\paragraph{Reasoning Distillation.} Parameter-efficient fine-tuning (LoRA) was applied to Qwen2.5-7B-Instruct using LLaMA-Factory (32K input, 1K output) on 8xA100- 80G GPUs.

\paragraph{Nuclei-Conditioned Summarization.} Using distilled backbones $B=\{e_{k}^{\mathrm{nuclei}}\}$ as input, Qwen2.5-0.5B-Instruct was full fine-tuned on $\{B,S^{\mathrm{ref}}\}$.

\begin{table}[h]
	\centering
	\caption{Results on MovieSum: ROUGE (R$_1$, R$_2$, R$_L$), compression ratio (Comp.), and BERTScore (BS$_P$, BS$_R$, BS$_{F1}$).}
	\label{tab:results}
	\begin{tabular*}{\textwidth}{@{\extracolsep{\fill}}l|l|P{2}{2}|P{1}{2}|P{2}{2}|c|P{2}{2}|P{2}{2}|P{2}{2}}
		\hline
		\textbf{Type} & \textbf{Model} & \textbf{R$_1$} & \textbf{R$_2$} & \textbf{R$_L$} & \textbf{Comp.} & \textbf{BS$_P$} & \textbf{BS$_R$} & \textbf{BS$_{F1}$} \\
		\hline
		\multirow{8}{*}{$\uparrow$ \textbf{Extractive}} 
		& Lead-512      & 10.35 & 1.27 & 9.84  & /       & \textit{49.25} & \textit{43.59} & \textit{46.23} \\
		& Lead-768      & 14.43 & 1.79 & 13.76 & /       & \textit{49.29} & \textit{45.70} & \textit{47.41} \\
		& Lead-1024     & 17.93 & 2.24 & 17.15 & /       & \textit{49.12} & \textit{46.91} & \textit{47.98} \\
		& TextRank      & 33.32 & 5.27 & 32.10 & /       & \textit{51.46} & \textit{52.47} & \textit{51.85} \\
		& FLAN-UL2      & 23.62 & 4.29 & 22.01 & \textbf{27.6\%}  & \textit{52.90} & \textit{49.57} & \textit{50.87} \\
		& Vicuna        & 16.35 & 3.55 & 15.44 & 55.2\%  & \textit{48.89} & \textit{48.49} & \textit{47.07} \\
		& TextRank+Vicuna& 17.14& 3.68 & 15.47 & 55.2\%  & \textit{59.24} & \textit{49.05} & \textit{53.57} \\
		& MW-Vicuna     & 19.56 & 3.32 & 18.57 & 55.2\%  & \textit{54.95} & \textit{48.70} & \textit{51.53} \\
		\hline
		\textbf{Hybrid}     
		& \textbf{S\textsuperscript{2}tory (Ours)} 
		& \textbf{45.98} & \textbf{7.93} & \textbf{42.45} & 28.4\%  & 59.18 & \textbf{59.36} & \textbf{59.23} \\
		\hline
		\multirow{6}{*}{$\downarrow$ \textbf{Abstractive}}
		& LED-Desc.     & \textit{44.72} & \textit{9.72} & \textit{42.92} & 55.2\% & \textbf{59.47} & 58.45 & 58.92 \\
		& LED-Heur.     & \textit{44.45} & \textit{9.78} & \textit{42.71} & 55.2\% & 58.93 & 58.15 & 58.54 \\
		& LED-Dialogue  & \textit{44.68} & \textit{10.02}& \textit{42.94} & 55.2\% & 59.30 & 58.29 & 58.74 \\
		& LED           & \textit{44.85} & \textit{9.83} & \textit{43.12} & 55.2\% & 59.11 & 58.43 & 58.73 \\
		& LongT5        & \textit{41.49} & \textit{8.54} & \textit{39.78} & 55.2\% & 56.09 & 55.36 & 55.68 \\
		& Pegasus-X     & \textit{42.42} & \textit{8.16} & \textit{40.63} & 55.2\% & 58.81 & 50.56 & 54.36 \\
		\hline
	\end{tabular*}
\end{table}

\subsection{Dataset and Baselines}
\paragraph{Dataset.} We focus on movie screenplays, which feature XML-style structural formatting. Our experiments use the MovieSum dataset~\cite{saxena2024moviesum}, a near-complete superset of existing screenplay corpora. For example, 98\% of the MENSA test set~\cite{saxena2024select} is contained within MovieSum. To assess generalization beyond screenplay-style narratives, we also evaluate on BookSum~\cite{kryscinski2022booksum}, a long-form prose corpus without XML formatting, demonstrating that NEAgent transfers through narratological reasoning rather than dataset-specific pattern learning.

\paragraph{Baselines.} To ensure comprehensive evaluation, we compare against a diverse set of baselines. These include extractive techniques (e.g., TextRank~\cite{mihalcea2004textrank}), instructiontuned LLMs (e.g., Vicuna~\cite{zheng2023judging}, FLAN-UL2~\cite{tay2022ul2}), and long-context summarization models (e.g., Pegasus-X~\cite{phang2023investigating}, LongT5~\cite{guo2022longt5}, LED~\cite{beltagy2020longformer}). All results are either adopted from prior work or reproduced under the same evaluation protocol as previous studies to ensure fair comparison and consistent evaluation metrics.

\subsection{Main Results}
Our experiments on the MovieSum benchmark show that S$^{2}$tory effectively combines the strengths of extractive and abstractive summarization. As shown in Table 1, our model achieves exceptional ROUGE scores $(\mathrm{R}_{1}=45.98, \mathrm{R}_{2}=7.93, \mathrm{R}_{L}=42.45)$, outperforming all extractive baselines by 32-38\% while matching the $\mathrm{R}_{L}$ performance of abstractive models.

S$^{2}$tory also achieves the highest BERTScore recall (59.36) among all methods, indicating superior semantic fidelity to reference summaries. Crucially, it accomplishes this with a compression ratio of only 28.4\%---less than half that of abstractive methods (55.2\%)---demonstrating its ability to generate concise yet comprehensive summaries of extremely long movie scripts.

These results validate our approach: by strategically integrating extractive precision with abstractive flexibility, S$^{2}$tory achieves an optimal balance between information coverage, semantic quality, and conciseness that neither purely extractive nor purely abstractive methods can match for long-form content summarization.

\section{Further Analysis}
We further analyze the model from four perspectives: ablation, human evaluation, qualitative visualization, and out-of-domain generalization.

\subsection{Ablation Study}
To assess the impact of character trajectory modeling on nucleus identification, we conduct an ablation study where the trajectory-based profiling module is removed from NEAgent. In the full model, NEAgent constructs structured profiles for each character, capturing evolving goals, identities, and inter-character dependencies, which serve as the foundation for evaluating narrative indispensability. Without this module, the agent directly applies nucleus-satellite reasoning on raw text, losing access to causal continuity in character development.

As shown in Table~\ref{tab:ablation_study}, removing trajectory modeling results in a marked drop in performance, with BERTScore F1 decreasing from 59.23 to 53.69. This confirms that modeling character development trajectories provides essential structural grounding for identifying narrative nuclei and maintaining cross-scene coherence in long-form scripts.

\begin{table}[h]
	\centering
	\caption{Ablation study on character trajectory modeling in nucleus reasoning.}
	\label{tab:ablation_study}
	\begin{tabular}{c|c|c|c}
		\hline
		Method Variant & BertScore-P & BertScore-R & BertScore-F1 \\
		\hline
		NEAgent w/o trajectory profiling & 53.09 & 55.28 & 53.69 \\
		NEAgent w/ trajectory profiling & \textbf{59.18} & \textbf{59.36} & \textbf{59.23} \\
		\hline
	\end{tabular}
\end{table}

\subsection{Human Evaluation}
Although the extracted nuclei are intermediate outputs, their quality directly affects summarization performance. We conducted a human evaluation along four narratological dimensions: \textit{indispensability}, \textit{coherence}, \textit{character consistency}, and \textit{satellite reduction}. Each was rated on a five-point scale (1-5) by doctoral students trained in narrative theory. The full evaluation rubric is provided in Appendix~\ref{app:human_eval}.

\begin{table}[h]
	\centering
	\caption{Evaluation results across four narrative dimensions (1-5 scale).}
	\label{tab:narrative_dimensions}
	\begin{tabular}{c|c|c}
		\hline
		Dimension & Auto Metric & Human Metric \\
		\hline
		Indispensability & 3.59 & 3.84 \\
		Coherence & 3.79 & 3.91 \\
		Character Consistency & 3.97 & 4.18 \\
		Satellite Reduction & 3.41 & 3.83 \\
		\hline
	\end{tabular}
\end{table}

As shown in Table~\ref{tab:narrative_dimensions}, human ratings consistently exceed automatic scores, particularly in \textit{satellite reduction}, where GPT-4o-mini tends to overvalue descriptive or emotional details. This gap indicates that large models capture surface fluency but struggle with the functional segmentation that defines narrative structure, whereas our nuclei extraction aligns more closely with narratological judgments.

\subsection{Out-of-Domain Generalization}
To evaluate cross-domain robustness, we directly apply the 7B distilled model to the BookSum corpus without further tuning. BookSum preserves long-form narrative coherence but lacks screenplay formatting, making it suitable for testing whether the model generalizes through narratological reasoning rather than XML-specific cues.

Because BookSum has no annotated nuclei, we adopt an LLM-as-Judge protocol~\cite{zheng2023judging}, prompting multiple LLMs to assess whether each generated nucleus constitutes a structurally essential event. Each case is labeled as positive or negative, while rejected indicates that the LLM declined to respond due to policy or copyright constraints.

\begin{table}[h]
	\centering
	\caption{Out-of-domain evaluation on BookSum using the LLM-as-Judge protocol.}
	\label{tab:ood_evaluation}
	\begin{tabular}{c|c|c|c}
		\hline
		Evaluator & Positive (\%) & Negative (\%) & Rejected (\%) \\
		\hline
		GPT-4.1 & 92.45 & 5.45 & 2.10 \\
		Qwen3-235B-A22B & 78.34 & 21.24 & 0.42 \\
		DeepSeek-R1-671B & 84.71 & 15.55 & 0.28 \\
		\hline
		\textbf{Average} & \textbf{85.17} & \textbf{14.08} & \textbf{0.93} \\
		\hline
	\end{tabular}
\end{table}

As shown in Table~\ref{tab:ood_evaluation}, over 85\% of the generated nuclei are judged as narratively essential by large evaluators, demonstrating strong cross-domain generalization. This suggests that NEAgent internalizes character-centric causal reasoning rather than overfitting to screenplay structures, enabling consistent narrative interpretation across diverse text domains.

\subsection{Case Study}
\begin{figure}[ht]
	\centering
	\begin{minipage}[b]{0.44\textwidth}
		\centering
		\includegraphics[width=\textwidth]{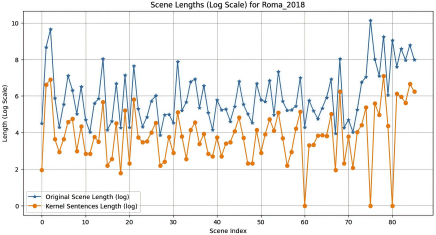}
		\captionof{figure}{Scene-length distribution: Original(Blue) vs. Nuclei(Orange).}
		\label{fig:scene_length}
	\end{minipage}%
	\hfill
	\begin{minipage}[b]{0.51\textwidth}
		\centering
		\includegraphics[width=\textwidth]{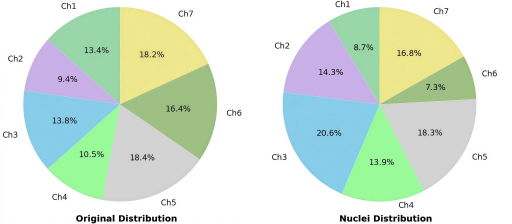}
		\captionof{figure}{Chapter-wise token distribution: Original(Left) vs. Nuclei(Right).}
		\label{fig:token_distribution}
	\end{minipage}
\end{figure}

We present a qualitative case study on \textit{Roma(2018)} to illustrate how S$^{2}$tory preserves narrative rhythm while achieving substantial compression. As shown in Fig~\ref{fig:scene_length}, the logarithmic scene-length distributions of the original screenplay (blue) and the generated nuclei (orange) exhibit a highly aligned oscillatory pattern across 85 scenes. Despite significant token reduction, the temporal fluctuations in narrative density are faithfully preserved, indicating that S$^{2}$tory maintains the underlying rhythmic structure of the film.

This alignment is not coincidental: key emotional or structural peaks -- such as the sharp length spike in Scene 79 -- are preserved in both versions, reflecting principled modeling of cinematic pacing. Moreover, as demonstrated in the chapterwise distribution for Book-151 (right panel), the generated nuclei maintain a structurally coherent proportionality across chapters. For instance, Fig~\ref{fig:token_distribution} increases in relative share from 13.8\% to 20.6\%, suggesting targeted condensation of less salient content while preserving the prominence of core narrative arcs.

Together, these results show that S$^{2}$tory does not compress via uniform truncation, but by selectively preserving the narrative pulse -- capturing both microlevel rhythmic dynamics and macro-level structural emphasis. This dual fidelity ensures that the distilled narrative remains faithful to the original in terms of dramatic tension and narrative cadence.


\section{Conclusion}
We presented S$^{2}$tory, a narratology-grounded framework for screenplay summarization. A Narrative Expert Agent (NEAgent) reasons over character trajectories to isolate plot nuclei from satellites, with its reasoning distilled into a compact model that conditions an abstractive summarizer. This demonstrates that narratology-guided reasoning provides a principled foundation for long-form story understanding. Future work will refine the causal link between nuclei and character state changes and extend the framework beyond screenplay conventions.

\appendix 
\section{Appendix}

\subsection{Prompt Details}
\label{app:prompt_detail}
Due to page constraints, we present a prompt card summarizing key components from our full prompt set. All narrative theories are implemented through carefully designed prompts to guide the NEAgent's reasoning process.

\textbf{1. Pronoun Replacement.} Expert linguistics role; identifies coreference clusters with character indices; groups mentions referring to same entities.

\textbf{2. Entity Profile Update.} Expert information extraction role; updates structured entity records using coreference clusters; maintains field merge rules.

\textbf{3. Narrative Units.} Replaces pronouns for independent readability; preserves sentence structure; segments by standard punctuation.

\textbf{4. Counterfactual Analysis.} Evaluates script coherence when removing sentences; assesses 5 dimensions including key information preservation.

\textbf{5. Kernel Events.} Based on Barthes' narrative theory; extracts only essential plot-driving elements; outputs micro-drama text ending with STOP token.

\textbf{6. Voting Protocol.} LLM-as-judge voting mechanism; selects most consistent events across multiple extraction attempts; resolves conflicts through majority rule.

\textbf{7. OOD Verification.} Validates drama miniaturization task; judges if compressed screenplay preserves narrative essence under extreme length constraints.

\subsection{Human Evaluation Criteria}
\label{app:human_eval} 
Human annotators assessed the output quality based on the following criteria:

\textbf{1. Indispensability (Mainline Necessity).} 1=irrelevant to main plot; 2=partly related, missing causal links; 3=mostly covers plot, minor gaps; 4=complete and logical; 5=precise and non-redundant.

\textbf{2. Coherence.} 1=fragmented; 2=weak logic, disjoint; 3=generally coherent; 4=smooth and consistent; 5=fully coherent and well-structured.

\textbf{3. Character Consistency.} 1=irrational or contradictory; 2=partly consistent, major jumps; 3=mostly consistent; 4=logical overall; 5=fully consistent with growth.

\textbf{4. Satellite Reduction.} 1=mostly redundant satellites; 2=over 50\% satellites; 3=30-40\% satellites; 4=10-20\% satellites; 5=nearly none, pure mainline.

%
%
%
\bibliographystyle{splncs04}
\bibliography{mybibliography}
%
%
%
%
%
\end{document}